\documentclass[10pt,twocolumn,letterpaper]{article}

\usepackage{cvpr}
\usepackage{times}
\usepackage{epsfig}
\usepackage{graphicx}
\usepackage{amsmath}
\usepackage{amssymb}
\usepackage{subfigure}
\usepackage{multirow}
\usepackage{xcolor}

\usepackage[pagebackref=true,breaklinks=true,letterpaper=true,colorlinks,bookmarks=false]{hyperref}

\cvprfinalcopy % *** Uncomment this line for the final submission

\def\cvprPaperID{65} % *** Enter the CVPR Paper ID here

\ifcvprfinal\fi
\begin{document}

%%%%%%%%% TITLE
\title{Multi-scale deep neural networks for real image super-resolution}

\author{Shangqi Gao\\
Fudan University\\
Shanghai, China\\
{\tt\small 18110980005@fudan.edu.cn}
\and
Xiahai Zhuang\\
Fudan University\\
Shanghai, China\\
{\tt\small zxh@fudan.edu.cn}
}

\maketitle
%\thispagestyle{empty}

%%%%%%%%% ABSTRACT
\begin{abstract}
	Single image super-resolution (SR) is extremely difficult if the upscaling factors of image pairs are unknown and different from each other, which is common in real image SR. To tackle the difficulty, we develop two multi-scale deep neural networks (MsDNN) in this work. Firstly, due to the high computation complexity in high-resolution spaces, we process an input image mainly in two different downscaling spaces, which could greatly lower the usage of GPU memory. Then, to reconstruct the details of an image, we design a multi-scale residual network (MsRN) in the downscaling spaces based on the residual blocks. Besides, we propose a multi-scale dense network based on the dense blocks to compare with MsRN. Finally, our empirical experiments show the robustness of MsDNN for image SR when the upscaling factor is unknown. According to the preliminary results of NTIRE 2019 image SR challenge, our team (ZXHresearch@fudan) ranks 21-st among all participants. The implementation of MsDNN is released \url{https://github.com/shangqigao/gsq-image-SR}
   
\end{abstract}

%%%%%%%%% BODY TEXT
\section{Introduction}
%the development of learning-based SR methods
Single image super-resolution (SR) aims at estimating the mapping from low-resolution (LR) to high-resolution (LR) spaces\cite{super02,low01,joint03}. Interpolation is one of the most common methods in super-resolving an image without referring the priors of its ground truth. However, several works showed that it would decimate the details of an image \cite{image01,image02,exploit01}. Recently, the learning-based methods have been widely applied in image SR thanks to its robust ability of recovering details. The learning-based methods aim at estimating the end-to-end mapping from LR to HR image pairs by adopting some well-known deep neural networks (DNN). The methods could be classified into two categories according to the manners of processing LR images, i.e., the methods in LR and HR spaces. 

%the image SR in LR space
The learning-based methods in LR space are generally proposed to reconstruct the SR image of an input under the given upscaling factor \cite{enhanced01,real01,accelerate01,deep03,photo01,wide01,real02}. For a given upscaling factor, the results of NTIRE 2017 image SR challenge showed that the learning-based methods perform robust when the input images are noise free \cite{ntire01}. In reality, natural images are inevitably accompanied with unknown noise, i.e., the downscaling operators from HR to LR images are different. Therefore, the noisy image SR is extremely challenging, which was verified in the NTIRE 2018 image SR challenge \cite{ntire02}.

%the　image SR in HR space
The learning-based methods in HR space are also developed to approximate the mapping from LR to HR image pairs using networks \cite{image02,deep02,accurate01}, but their inputs have the same size with the corresponding outputs. If the upscaling factor is given, the input will be the interpolation of an LR image. In reality, both close-shot and long-shot could be existed in an image dataset. The upscaling factors of close-shot and long-shot should be different if we want to obtain the images with high quality. Therefore, we expect to develop a method which has generalization capability to different upscaling factors. 
%the difficulty of image SR with unknown upscale, 

The NTIRE 2019 image SR challenge\footnote{http://www.vision.ee.ethz.ch/ntire19/} provided a dataset in which the sizes of each LR and HR image pair are the same, i.e., the upscaling factors of LR images are unknown and possibly different. The training dataset is small due to the fact that there are only 60 image pairs provided. Besides, the dataset is composed of the images obtained by two different cameras. Overall, the downscaling operators from HR to LR images are different. Therefore, one difficulty of this challenge is to develop a method which performs robust in super-resolving the LR images obtained by different downscaling operators. Another challenge is to explore an approach with low computation complexity since the LR images are very large in the testing phase. To tackle the former, we devote to designing a deep neural network (DNN) thanks to its robust ability in reconstructing the details of an image. To solve the latter, we propose a multi-scale model to downscale the input images.

% our muti-scale method
In this work, we develop two multi-scale deep neural networks to solve the problems from NTIRE 2019 image SR challenge. For the learning-based methods developed in HR space, the number of filters in each convolutional layer should be small due to the large sizes of testing images. To increase the features maps of each layer, we downscale an input to obtain its downscaling, and process the input and two downscaling images in HR  space (DS0), downscaling $ \times 2 $ (DS2) and downscaling $ \times 4 $ (DS4) spaces, respectively. Overall, we develop two networks in downscaling spaces to recover the details of input images, i.e., multi-scale residual networks (MsRN) and multi-scale dense network (MsDN). MsRN is mainly developed in DS2 and DS4 while MsDN is exploited in each space. Our experiments show that the training and testing efficiency in downscaling spaces are much higher due to the fact that the size of a downscaling image is greatly decreased.

The rest of this work is organized as follows. Section \ref{sec:2} shows the related works about the residual and dense networks for image SR. We develop two multi-scale networks in Section \ref{sec:3}. Section \ref{sec:4} gives the details of training strategies. The description of our experiments is presented in Section \ref{sec:5}. We discuss the proposed method in Section \ref{sec:6} and conclude this work in Section \ref{sec:7}.

\section{Related works}\label{sec:2}
The aim of single image SR is to estimate the mapping from LR to HR image pairs. Recently, DNN has been widely applied in image SR due to its ability of simulating complex mappings. Dong \etal \cite{image02} first proposed to approximate the mapping from LR to HR image pairs using a three layers convolutional neural network. Since then, other architectures, such as RNN \cite{conv01, under01, recurrent01}, ResNet \cite{deep04}, and GAN \cite{gan01}, have been applied in image SR. 

\begin{figure}[htp]
	\centering
	\subfigure[ResNet]{\includegraphics[width=0.3\linewidth]{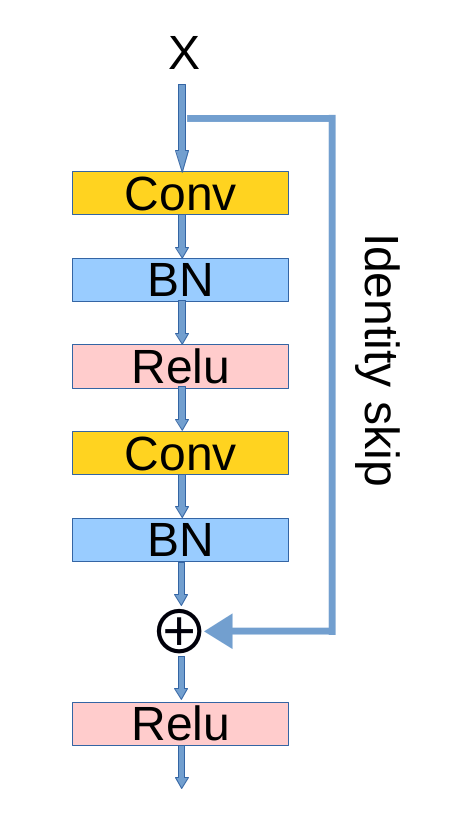}}
	\subfigure[SRResNet]{\includegraphics[width=0.3\linewidth]{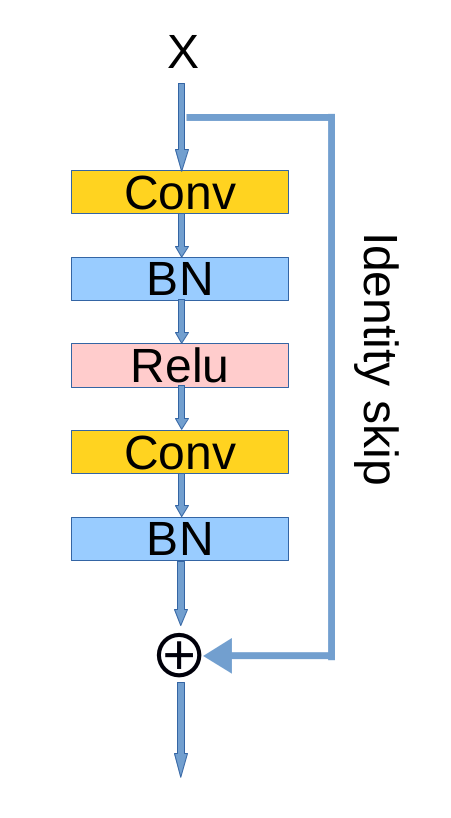}}
	\subfigure[EDSR]{\includegraphics[width=0.3\linewidth]{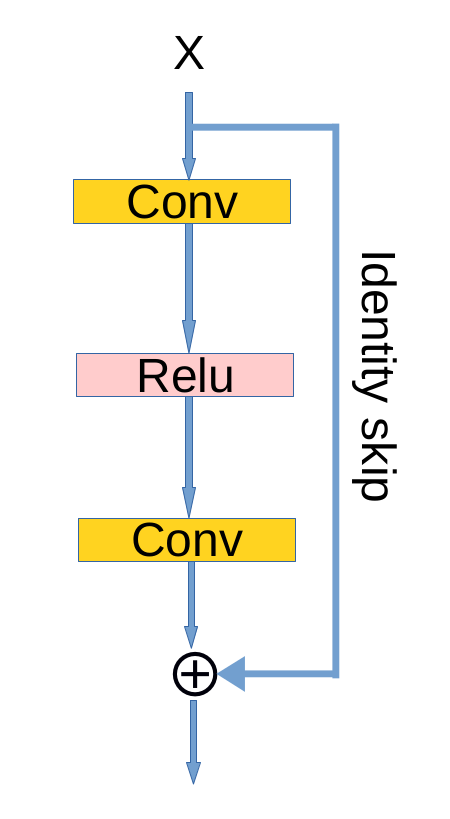}}
	\caption{The comparison of residual blocks in ResNet \cite{deep04}, SRResNet \cite{photo01}, and EDSR \cite{enhanced01}.}\label{fig:01}
\end{figure}

ResNet was first proposed by He \etal \cite{deep04} for the task of classification. Ledig \etal \cite{photo01} successfully introduce it to image SR and developed an approach referred as SRResNet. SRResNet preserved the batch normalization from original residual blocks, which was showed to consume amounts of memory in computation and restrict the range flexibility from networks for image deblurring \cite{deep05} and SR \cite{enhanced01}. Therefore, Lim \etal \cite{enhanced01} proposed a new residual block by removing the batch normalization and developed an enhanced deep residual network referred as EDSR. The comparison of ResNet, SRResNet and EDSR is showed in Figure \ref{fig:01}.

The skip connections in ResNet are critical since they constraint a residual block to learn the residual between its input and output \cite{deep04}. To explore the advantages of skip connection, DenseNet proposed to link all layers in the networks to efficiently train a very deep networks \cite{densenet}. To adopt the idea of DenseNet for image SR, Tong \etal \cite{image07} developed a method, referred as SRDenseNet, to estimate the mapping from LR to HR image pairs. Motivated by the idea of EDSR and SRDenseNet, we will design two multi-scale deep neural networks for single image SR with unknown upscaling factors and downscaling operators in Section \ref{sec:3}.

\section{Multi-scale deep neural networks}\label{sec:3}
In reality, images are often obtained by different kinds of cameras. Besides, to reconstruct the SR image with high quality, the upscaling factors of close-shot and long-shot could be different. Overall, the challenge of realistic image SR is that we need to develop a method which has generalization capacity to both the upscaling factors and downscaling operators.  

\subsection{Multi-scale residual networks}\label{sub:3.1}
Deep residual networks show a big potentiality in approximating the end-to-end mapping from LR to HR spaces. The results of NTIRE 2018 image SR challenge \cite{ntire02} showed that the ResNet-based methods perform robust when LR images are obtained by unknown downscaling operators.  However, for NTIRE 2019 image SR challenge, it is extremely expensive to develop a deep residual networks in LR space since the sizes of LR and HR image pair are the same. To be specific, the HR image from this challenge is firstly downscaled by an unknown operator, then the downscaling image is interpolated to obtain a LR image with the same size as the HR image. Therefore, the number of filters in each layer  will be limited by the single GPU memory due to the high computation complexity in LR space. 
\begin{figure}
	\centering
	\includegraphics[width=0.98\linewidth]{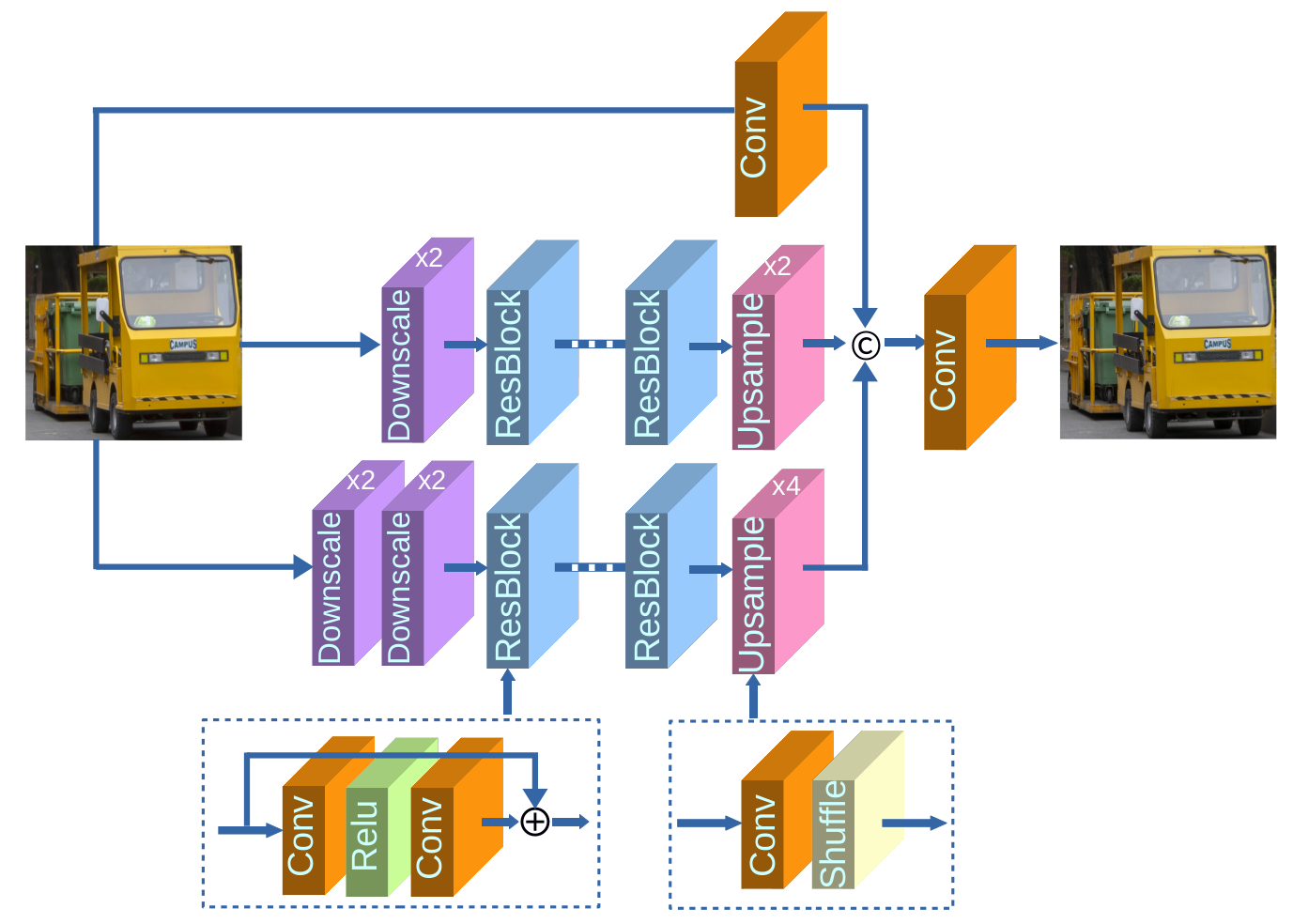}
	\caption{The architecture of the proposed multi-scale residual networks.}
	\label{fig:02}
\end{figure}

The previous works have verified that a wide convolutional layer could improve the performance of deep residual networks \cite{enhanced01, wide01}. Inspired by this, we propose a multi-scale residual network (MsRN) to get rid of the limitation of GPU memory and widen the convolutional layers of this network, as shown in Figure \ref{fig:02}. The architecture of MsRN in DS2 or DS4 space is a deep residual network, and in DS0 space is composed of a convolutional layer. Moreover, the \textit{Downscale} in Figure \ref{fig:02} denotes a convolutional layer with the strides equal to 2. There are several advantages about this architecture. First, the training and testing of MsRN are cheap due to the fact that the main computation is implemented in the two downscaling spaces. Besides, the maximal number of filters in DS2 could be four times of it in DS0, which is important to improve the performance of MsRN. Finally, the connection in DS0 could reduce the loss of information from an LR image in the process of downscaling.

\subsection{Multi-scale dense networks}\label{sec:3.2}
As shown in ResNet, the output of an residual block is the sum of its main and identity connections. Although the structure performs robust in learning the residual between its input and output, it may leads to the loss of useful information from the input. To solve the problem, we have designed a skip connection in DS0 space for MsRN, but it can not prevent the loss of information in DS2 and DS4 spaces which is induced by the depth of networks. 
\begin{figure}
	\centering
	\includegraphics[width=0.98\linewidth]{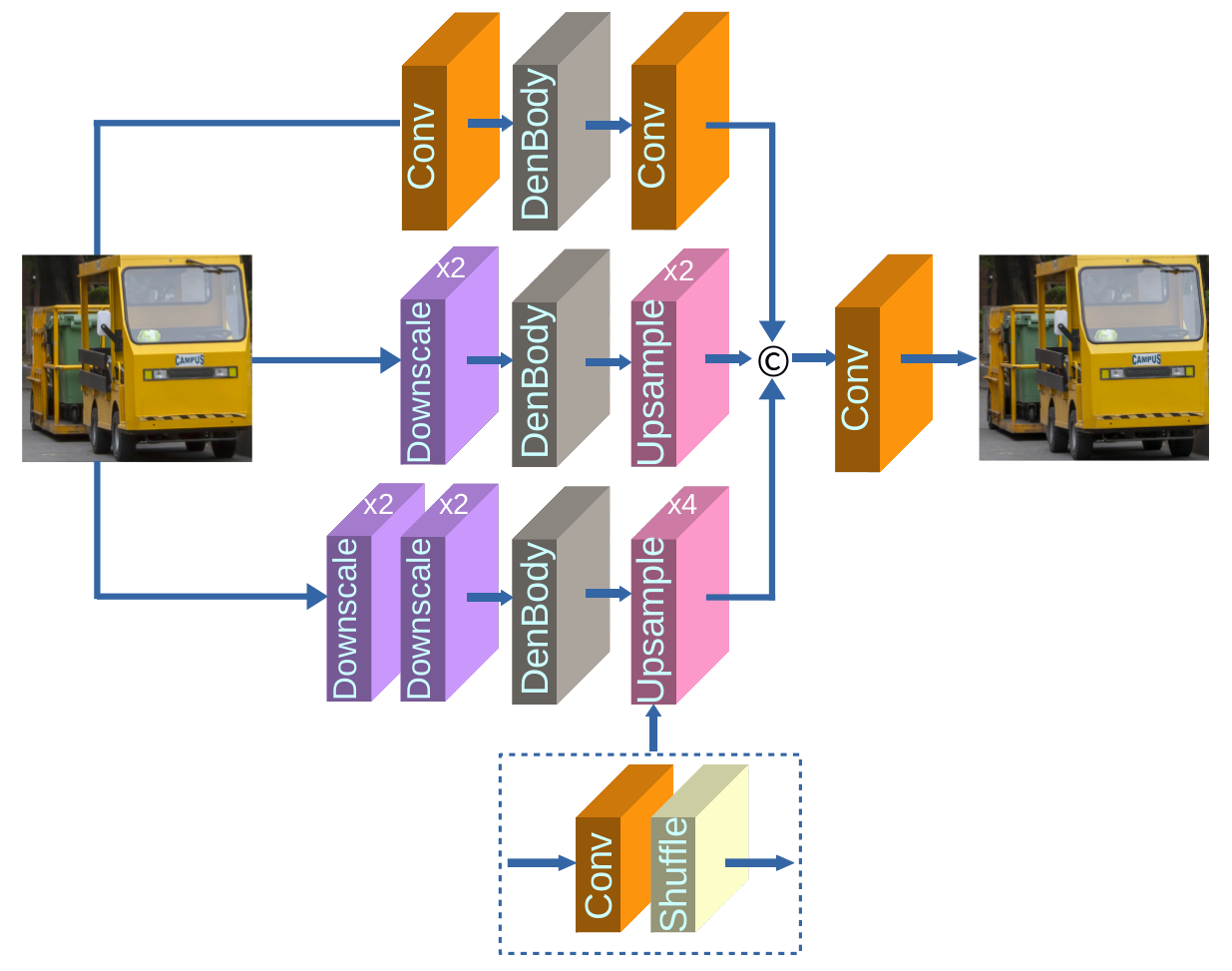}
	\caption{The architecture of the proposed multi-scale dense networks.}
	\label{fig:03}
\end{figure}

\begin{figure}
	\centering
	\includegraphics[width=0.98\linewidth]{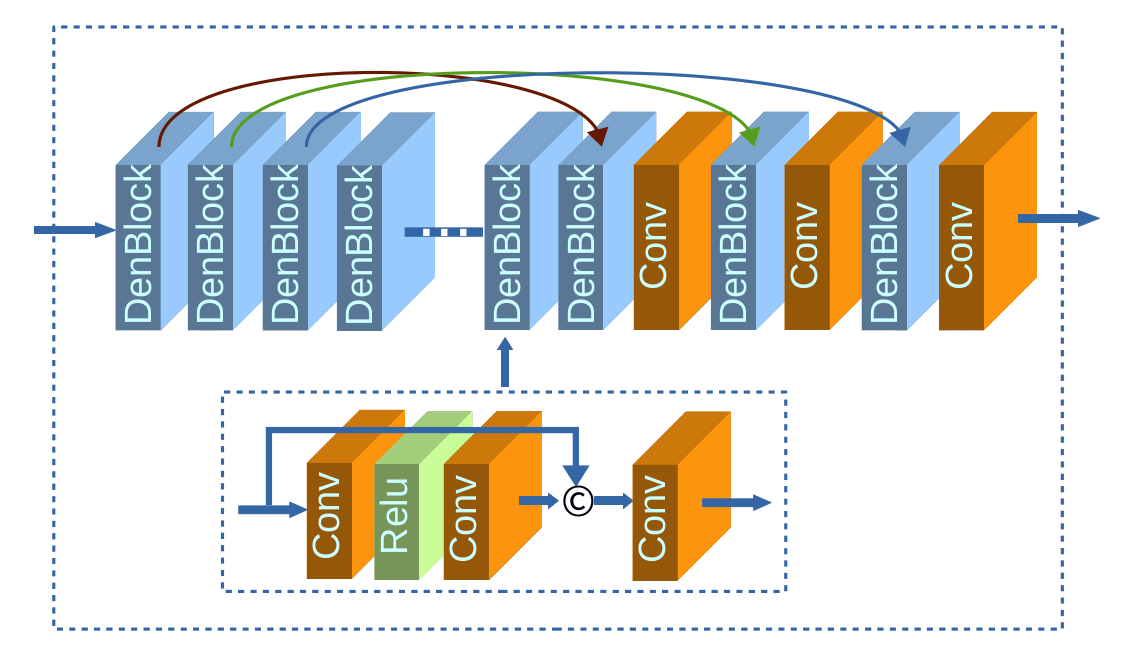}
	\caption{The detailed structure of \textit{DenBody} in the multi-scale dense networks.}
	\label{fig:04}
\end{figure}

The DenseNet is proposed to sufficiently take the advantages of skip connections. Motivated by this, we develop a multi-scale dense network (MsDN) for image SR, as shown in Figure \ref{fig:03}.  The networks in DS0, DS2, and DS4 are mainly composed of the \textit{DenBody}, of which the detailed architecture is presented in Figure \ref{fig:04}.  As shown in Figure \ref{fig:04}, the first three blocks are concatenated with the final three blocks to reduce the loss of information. Compared with MsRN, the depth of MsDN could be much smaller due to the amount of skip connections. Therefore, the training and testing of MsDN are cheaper than MsRN although there is a dense network in DS0.

\section{Training and testing strategies}\label{sec:4}
As well as we known, the training of DNN on a small dataset is very difficult due to the overfitting. One method to tackle the difficulty is to use regularization, such as weight decay, but it would increase the computation complexity. Another approach to solve the problem is to augment a small dataset by flipping and rotations, which is widely utilized in computer vision. 

\begin{table*}[htp]
	\centering
	\caption{The settings of multi-scale deep neural networks.}
	\label{tab:01}
	\setlength{\tabcolsep}{7.8mm}{
		\begin{tabular}{c|c|c|c|c}
			\hline 
			Options&  Baseline-R   &MsRN&  Baseline-D   &MsDN\\ 
			\hline 
			$ \# $Total blocks&  $  (32, -, -) $ &$ (0, 32, 32) $&  $ (12, -, -) $&$ (12, 12, 12) $\\ 
			$ \# $Filters&  $ (16, -, -) $&$ (3, 96, 96) $&  $ (16, -, -) $&$ (12, 48, 96) $\\ 
			$ \# $Downscale  &  --& $ \left\lbrace 96, (48, 96)\right\rbrace  $  &  --& $ \left\lbrace 48, (48, 96) \right\rbrace $\\
			$ \# $Additions&  32& 64&  0& 0\\ 
			$ \# $Concatenations&  0&1&  15&46\\ 
			$ \# $Parameters&  9.4K&113.3K&  5.8K&91.9K\\ 
			\hline 
	\end{tabular} }
\end{table*}
The NTIRE 2019 image SR challenge provided a training dataset with only 60 image pairs, which is not enough for the training of deep networks. Therefore, in the training stage. we use the strategy of augmentation to prevent our models from overfitting. To be specific, let $ f_{lr} $ and $ f_{ud} $ denote the left-right and up-down flipping, respectively, and $ r $ denotes the 90 degree rotations. Then every training image would be mapped by the three mappings one by one with a possibility of one in two. This strategy has showed to be useful in increasing the diversity of a training dataset.

For image SR, the SR image of an input could be enhanced by averaging the SR images derived from the transformations of the input \cite{seven}. Therefore, in the testing phase, we utilize the strategy to enhance the performance of image SR. Concretely, Let us denote an input LR image as $ I_{LR} $, and MsDNN as $ F $, then the SR image of $ I_{LR} $ will be obtained by averaging the elements in the following set,
\begin{gather}
\left\lbrace F(I_0), F(I_1), F(I_2), F(I_3)\right\rbrace  \nonumber\\
\cup \left\lbrace F(r(I_0)), F(r(I_1)), F(r(I_2)), F(r(I_3)) \right\rbrace 
\end{gather}
, where $ I_0 = I_{LR} $, $ I_1 = f_{lr}(I_{LR}) $, $ I_2 = f_{ud}(I_{LR}) $, and $ I_3 = f_{ud}(f_{lr}(I_{LR})) $.

\section{Experiments}\label{sec:5}

\subsection{Details of implementation}\label{sub:5.1}
The dataset from NTIRE 2019 image SR challenge is composed of 60 training images, 20 validation images, and 20 testing images. The LR images of this dataset have the same size with the corresponding HR images, i.e., a LR image is the interpolation of an unknown downscaling HR image. We will test the performance of our method on the validation dataset since the ground truth of testing dataset is not public.

\begin{figure*}[htp]
	\centering
	\begin{minipage}{0.318\linewidth}
		\includegraphics[width=1\linewidth]{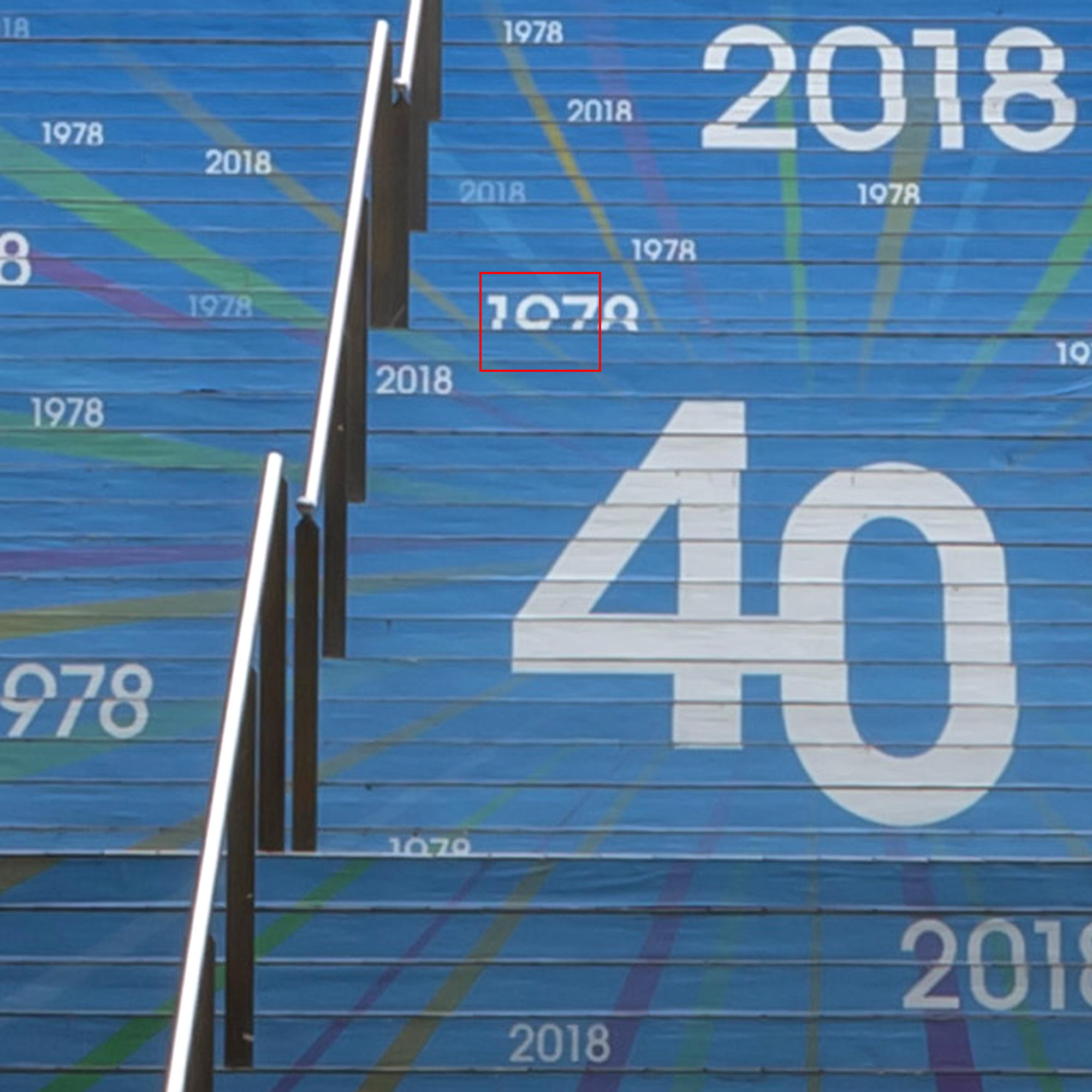}
		\centerline{Low-resolution image}
		\centerline{\textit{cam1\_08} from NTIRE 2019}
	\end{minipage}
	\begin{minipage}{0.16\linewidth}
		\includegraphics[width=1\linewidth]{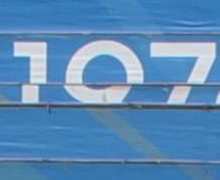}
		\centerline{HR}
		\centerline{(PSNR/SSIM)}
		\includegraphics[width=1\linewidth]{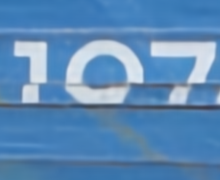}
		\centerline{MsRN-2}
		\centerline{(\color{blue}{29.62}/\color{blue}{0.8624})}
	\end{minipage}
	\begin{minipage}{0.16\linewidth}
		\includegraphics[width=1\linewidth]{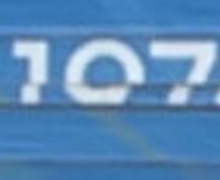}
		\centerline{LR}
		\centerline{(27.19/0.7919)}
		\includegraphics[width=1\linewidth]{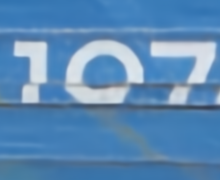}
		\centerline{MsRN-1}
		\centerline{(29.71/0.8636)}
	\end{minipage}
	\begin{minipage}{0.16\linewidth}
		\includegraphics[width=1\linewidth]{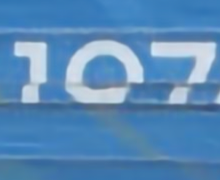}
		\centerline{Baseline-R}
		\centerline{(28.76/0.8374)}
		\includegraphics[width=1\linewidth]{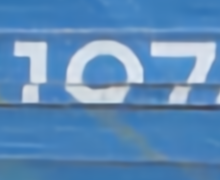}
		\centerline{MsDN-2}
		\centerline{(29.48/0.8586)}
	\end{minipage}
	\begin{minipage}{0.16\linewidth}
		\includegraphics[width=1\linewidth]{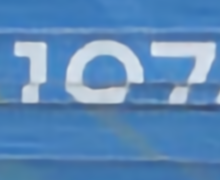}
		\centerline{Baseline-D}
		\centerline{(28.94/0.8466)}
		\includegraphics[width=1\linewidth]{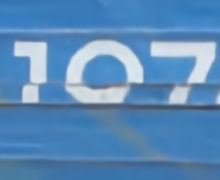}
		\centerline{MsDN-1}
		\centerline{(\color{red}{29.84/0.8669})}
	\end{minipage}
	\begin{minipage}{0.318\linewidth}
		\includegraphics[width=1\linewidth]{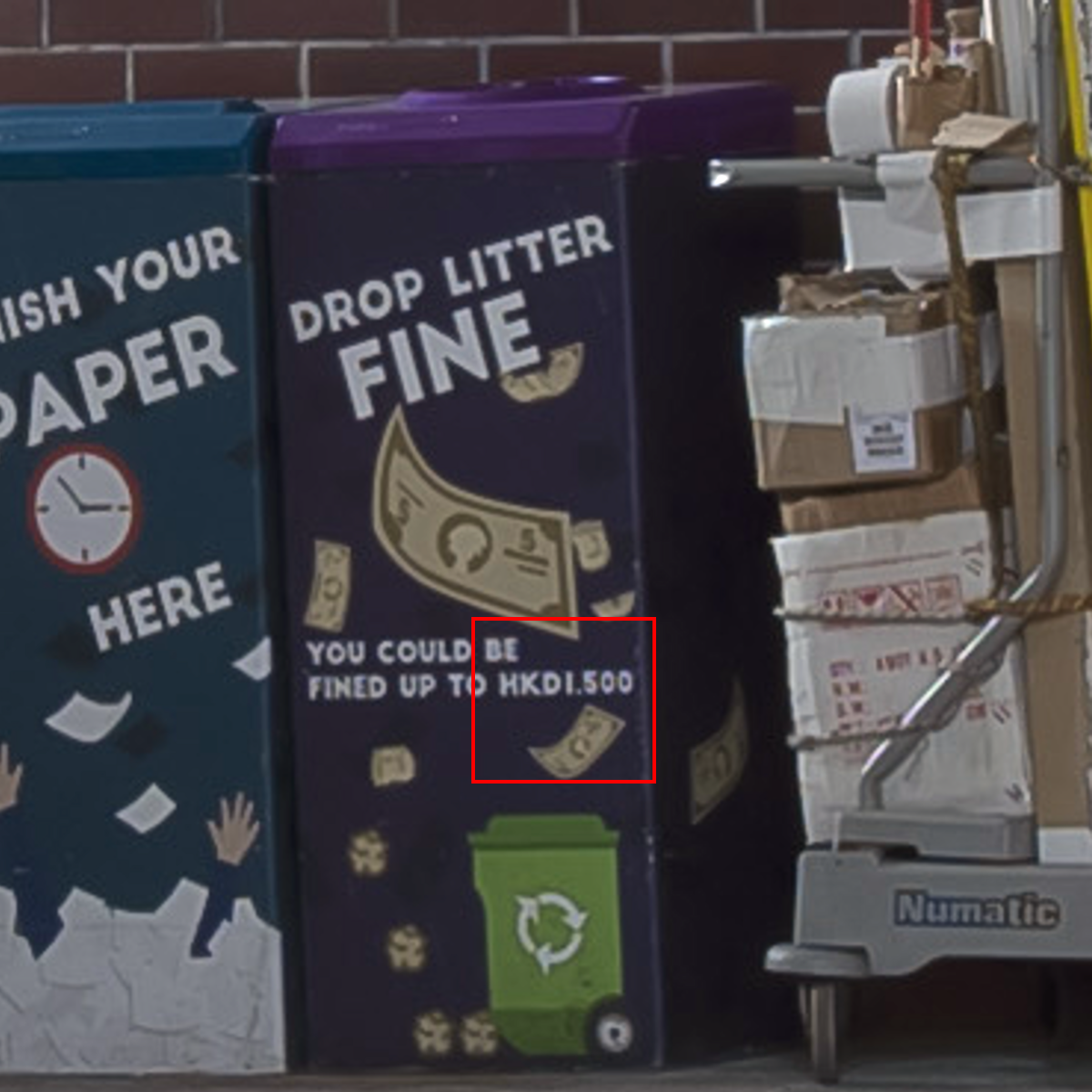}
		\centerline{Low-resolution image}
		\centerline{\textit{cam2\_02} from NTIRE 2019}
	\end{minipage}
	\begin{minipage}{0.16\linewidth}
		\includegraphics[width=1\linewidth]{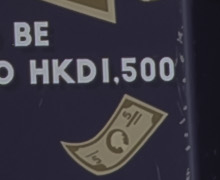}
		\centerline{HR}
		\centerline{(PSNR/SSIM)}
		\includegraphics[width=1\linewidth]{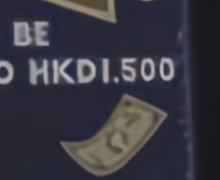}
		\centerline{MsRN-2}
		\centerline{(\color{blue}{30.77/0.9017})}
	\end{minipage}
	\begin{minipage}{0.16\linewidth}
		\includegraphics[width=1\linewidth]{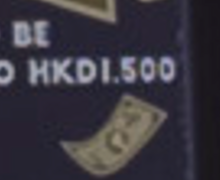}
		\centerline{LR}
		\centerline{(28.74/0.8516)}
		\includegraphics[width=1\linewidth]{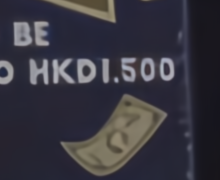}
		\centerline{MsRN-1}
		\centerline{(30.96/0.9042)}
	\end{minipage}
	\begin{minipage}{0.16\linewidth}
		\includegraphics[width=1\linewidth]{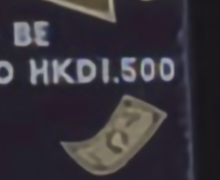}
		\centerline{Baseline-R}
		\centerline{(30.01/0.8897)}
		\includegraphics[width=1\linewidth]{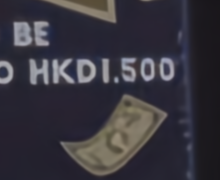}
		\centerline{MsDN-2}
		\centerline{(30.70/0.9016)}
	\end{minipage}
	\begin{minipage}{0.16\linewidth}
		\includegraphics[width=1\linewidth]{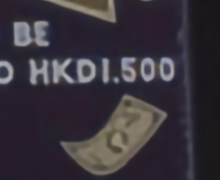}
		\centerline{Baseline-D}
		\centerline{(30.04/0.8929)}
		\includegraphics[width=1\linewidth]{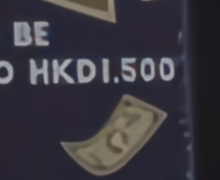}
		\centerline{MsDN-1}
		\centerline{(\color{red}{31.02/0.9054})}
	\end{minipage}
	
	\begin{minipage}{0.318\linewidth}
		\includegraphics[width=1\linewidth]{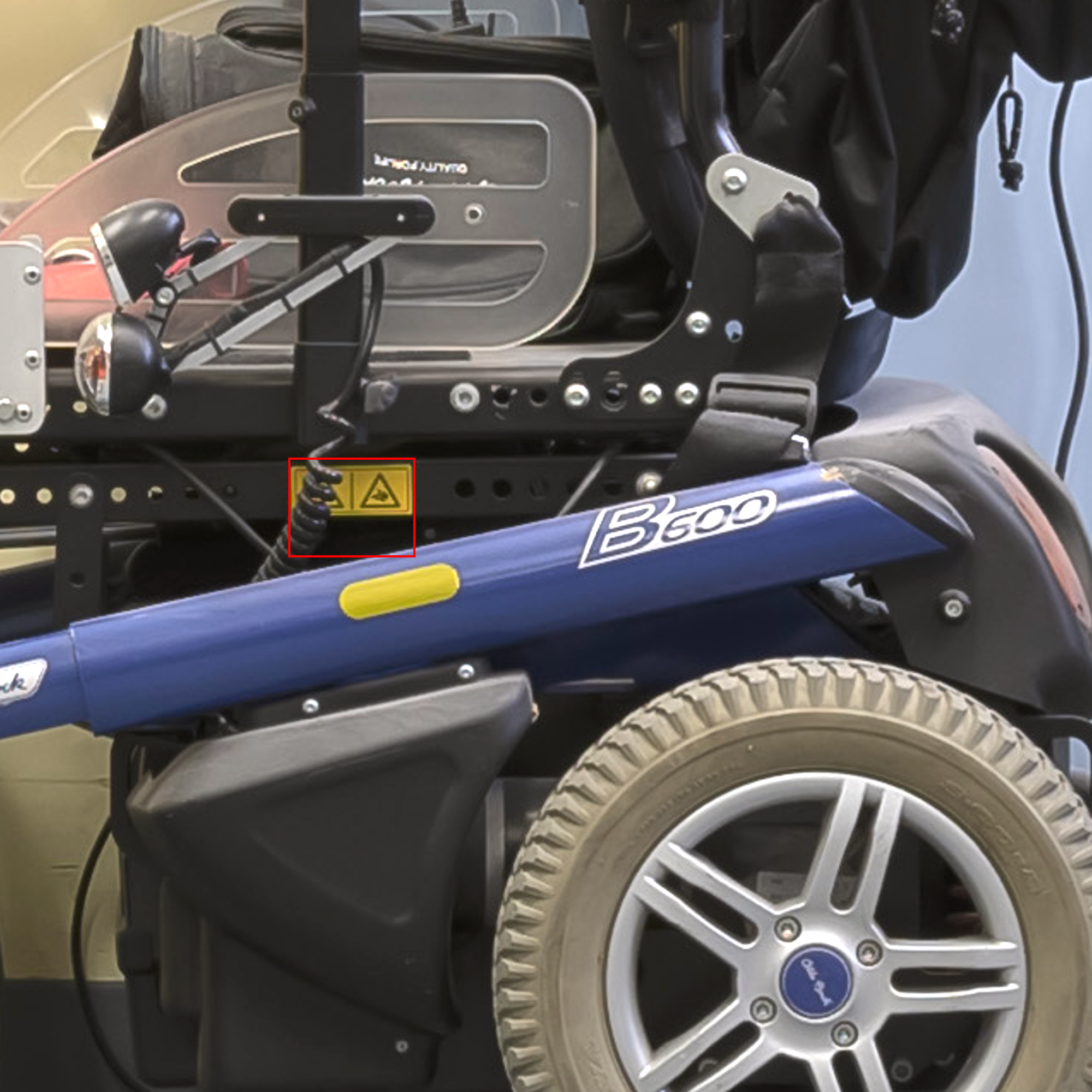}
		\centerline{Low-resolution image}
		\centerline{\textit{cam2\_06} from NTIRE 2019}
	\end{minipage}
	\begin{minipage}{0.16\linewidth}
		\includegraphics[width=1\linewidth]{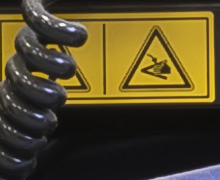}
		\centerline{HR}
		\centerline{(PSNR/SSIM)}
		\includegraphics[width=1\linewidth]{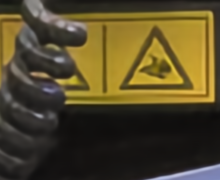}
		\centerline{MsRN-2}
		\centerline{(28.62/0.8697)}
	\end{minipage}
	\begin{minipage}{0.16\linewidth}
		\includegraphics[width=1\linewidth]{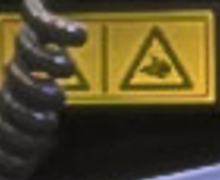}
		\centerline{LR}
		\centerline{(27.42/0.8416)}
		\includegraphics[width=1\linewidth]{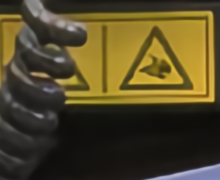}
		\centerline{MsRN-1}
		\centerline{(28.61/0.8723)}
	\end{minipage}
	\begin{minipage}{0.16\linewidth}
		\includegraphics[width=1\linewidth]{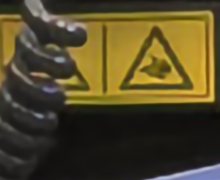}
		\centerline{Baseline-R}
		\centerline{(28.53/0.8666)}
		\includegraphics[width=1\linewidth]{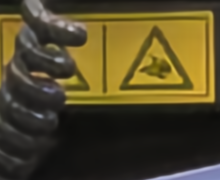}
		\centerline{MsDN-2}
		\centerline{(\color{blue}{28.70/0.8740})}
	\end{minipage}
	\begin{minipage}{0.16\linewidth}
		\includegraphics[width=1\linewidth]{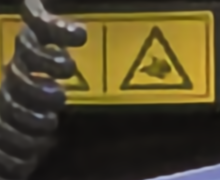}
		\centerline{Baseline-D}
		\centerline{(28.54/0.8678)}
		\includegraphics[width=1\linewidth]{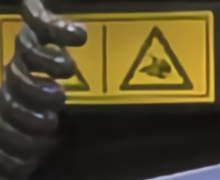}
		\centerline{MsDN-1}
		\centerline{(\color{red}{28.77/0.8728})}
	\end{minipage}
	\caption{The visualization of SR images which are reconstructed by MsRN and MsDN on the three images from validation dataset. The blue value denotes the maximum among the values obtained by $ L_2 $ norm based models. The red value denotes the maximum among the values obtained by $ L_1 $ norm based models.  }
	\label{fig:07}
\end{figure*}
There is a data pre-processing before training. To be specific, we randomly crop $ 64\times 64 $ pixels patches from LR images and augment them via the training strategies in Section \ref{fig:04}. Then these patches are shuffled and utilized to assemble the batches of size 16. Finally, the batches are used to train all models. Unless otherwise stated, the kernel size of convolutions in MsDNN is $ 3\times 3 $. To demonstrate the advantages of MsDNN, we set two baselines for MsRN and MsDN. Concretely, we design a deep residual network in DS0 space as the baseline of MsRN, which is referred as Baseline-R. In the same way, a deep dense network is developed in DS0 to be the baseline of MsDN, which is referred as Baseline-D.  The detailed settings of baselines, MsRN and MsDN are showed in Table \ref{tab:01}. Here, $ \# $Total blocks denotes the number of blocks in DS0, DS2, and DS4 spaces, respectively.  $ \# $Filters denotes the number of filters of a convolutional layer in DS0, DS2, and DS4 spaces, respectively. \textit{Downscale} in MsDNN is composed of a convolutional layer with the strides equal to 2. Therefore, the first element of $ \# $Downscale in Table \ref{tab:01} denotes the number of filters of \textit{Downscale} in DS2 space, and the second element denotes the number of filters of two \textit{Downscale} in DS4 space. $ \# $Additions and $ \# $Concatenations denote the number of addition and concatenation existed in MsDNN, i.e., the amount of \textcircled{+} and \textcircled{c}, respectively. 

For training, each model is trained by optimizing the $ L_2 $ or $ L_1 $ norm. All methods are trained by the ADAM optimizer, and settings of parameters are $ \beta_1 = 0.9 $, $ \beta_2 = 0.999 $, and $ \epsilon = 1\times 10^{-8} $. The training of each model is up to 100 million updates. The initial learning rate is $ 1\times 10^{-4} $, it decreases to 20 percent every 10 million updates when the updates are greater than 60 million. Finally, We implement our networks with TensorFlow and train our models on a device with 40 Intel Xeon 2.20 Ghz CPUs and 4 DTX 1080 Ti GPUs. The training of MsRN and MsDN cost about 30 and 20 hours on a single GPU, respectively. 

During testing, we utilize the testing strategy of enhancement in Section \ref{sec:4} to improve the performance of models on testing dataset. To evaluate the performance of our method, the standard Peak Signal To Noise Ratio (PSNR) and the Structural Similarity (SSIM) index are used. We could directly compute the PSNR and SSIM of a LR- and SR image pair since their sizes are the same. Finally, we show the quantitative comparisons of our models via the standard criteria.

\subsection{Comparisons of $ L_1 $ and $ L_2 $ losses}\label{sub:5.2}
The loss functions is critical to the training of networks due to the fact that each parameter of the networks is updated by optimizing the given losses. To compare the performance of MsDNN under different loss functions, we have trained MsDNN by optimizing $ L_1 $ and $ L_2 $ norm.  Concretely, we first obtain two models, i.e., MsRN-1 and MsRN-2, by training MsRN with $ L_1 $ and $ L_2 $. Then compute the PSNR and SSIM of two models in the training phase, to plot the convergent curves of MsRN-1 and MsRN-2 on a cropped validation dataset. We repeat the above process for MsDN to obtain two models, i.e., MsDN-1 and MsDN-2, and plot the convergent curves of these two models.
\begin{figure*}[htp]
	\centering
	\subfigure[PSNR of MsRN]{\includegraphics[width=0.48\linewidth]{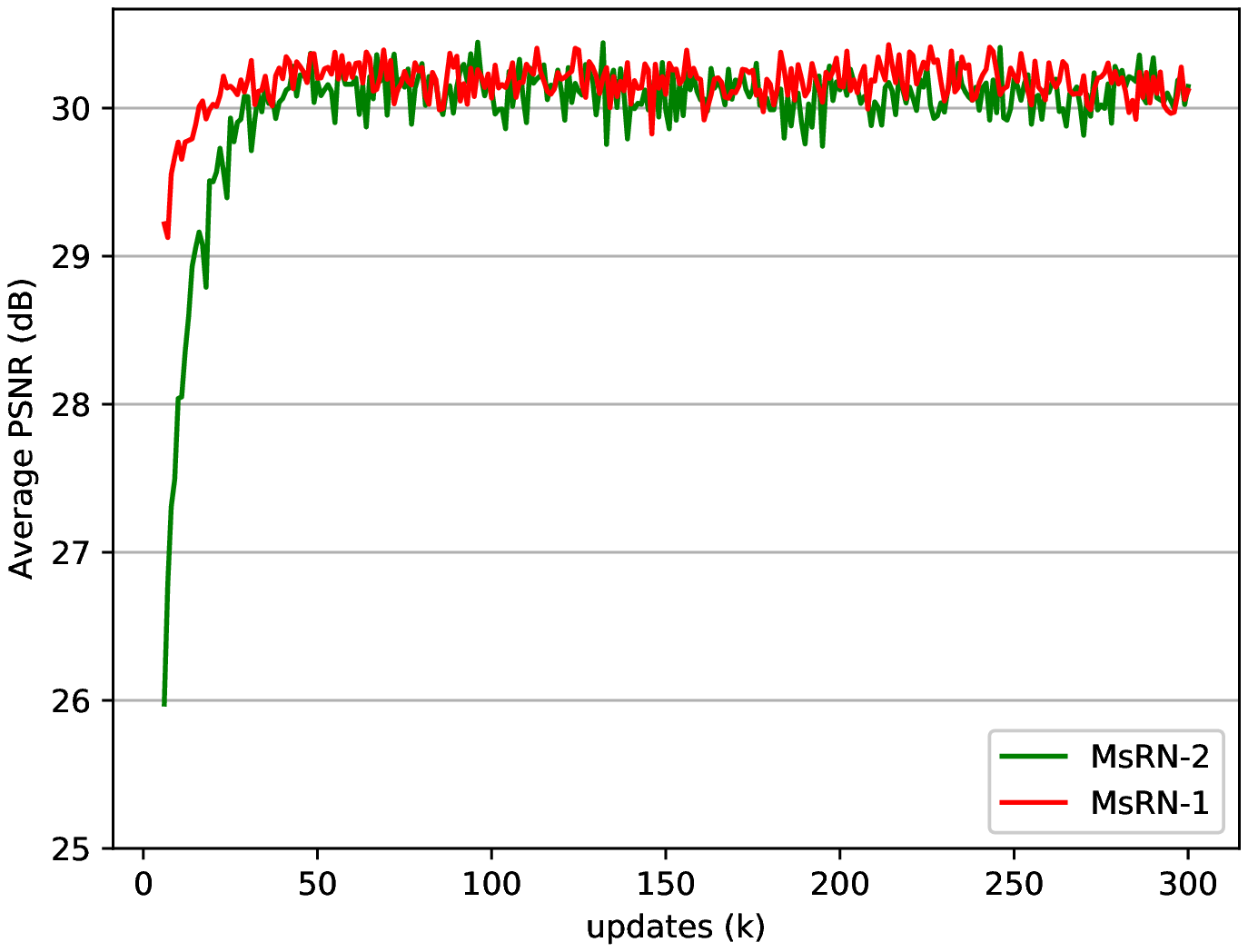}}
	\subfigure[SSIM of MsRN]{\includegraphics[width=0.48\linewidth]{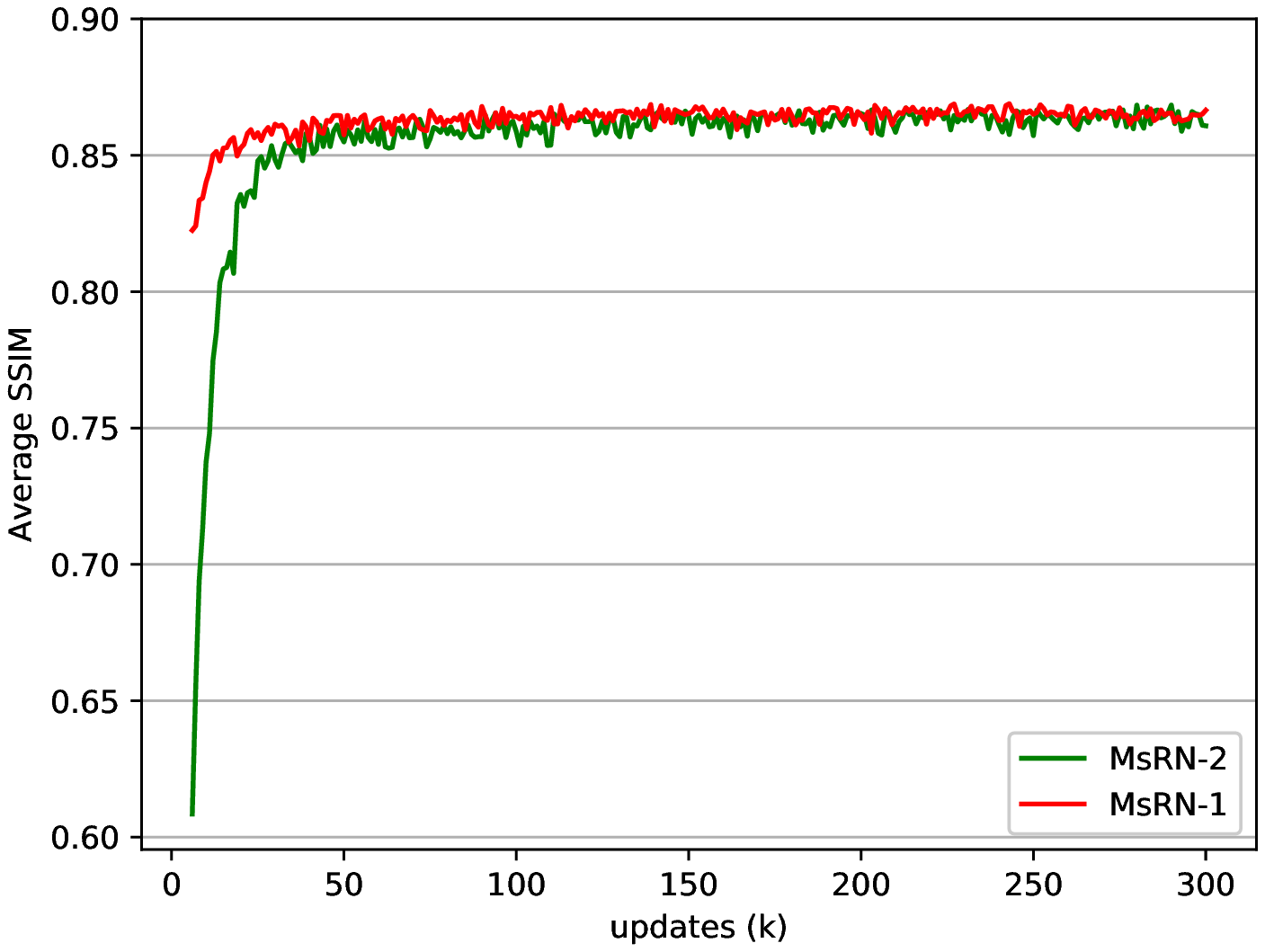}}
	\caption{The convergent curves of average PNSR and SSIM of MsRN on a cropped validation dataset in the training phase.}
	\label{fig:05}
\end{figure*}

\begin{figure*}[htp]
	\centering
	\subfigure[PSNR of MsDN]{\includegraphics[width=0.48\linewidth]{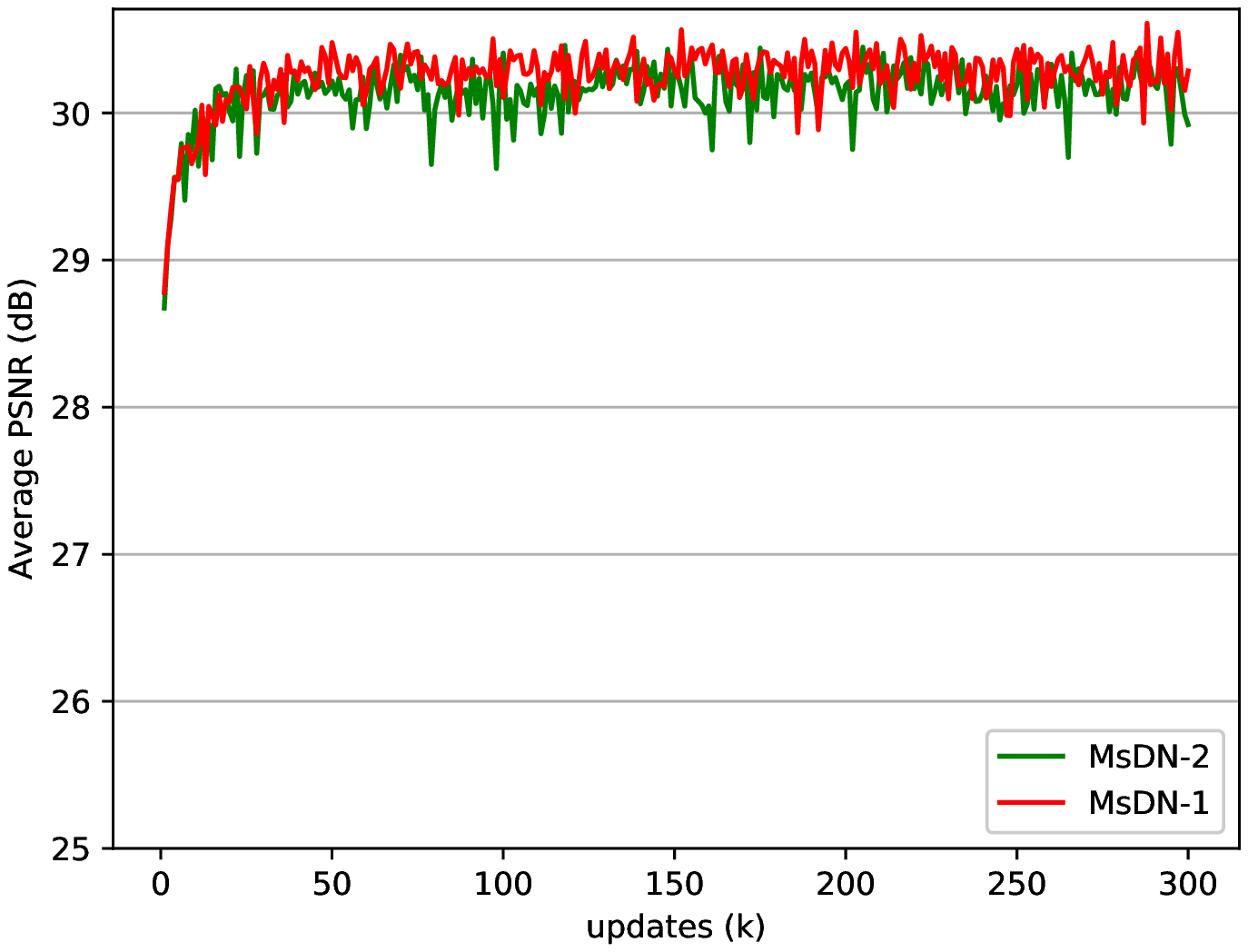}}
	\subfigure[SSIM of MsDN]{\includegraphics[width=0.48\linewidth]{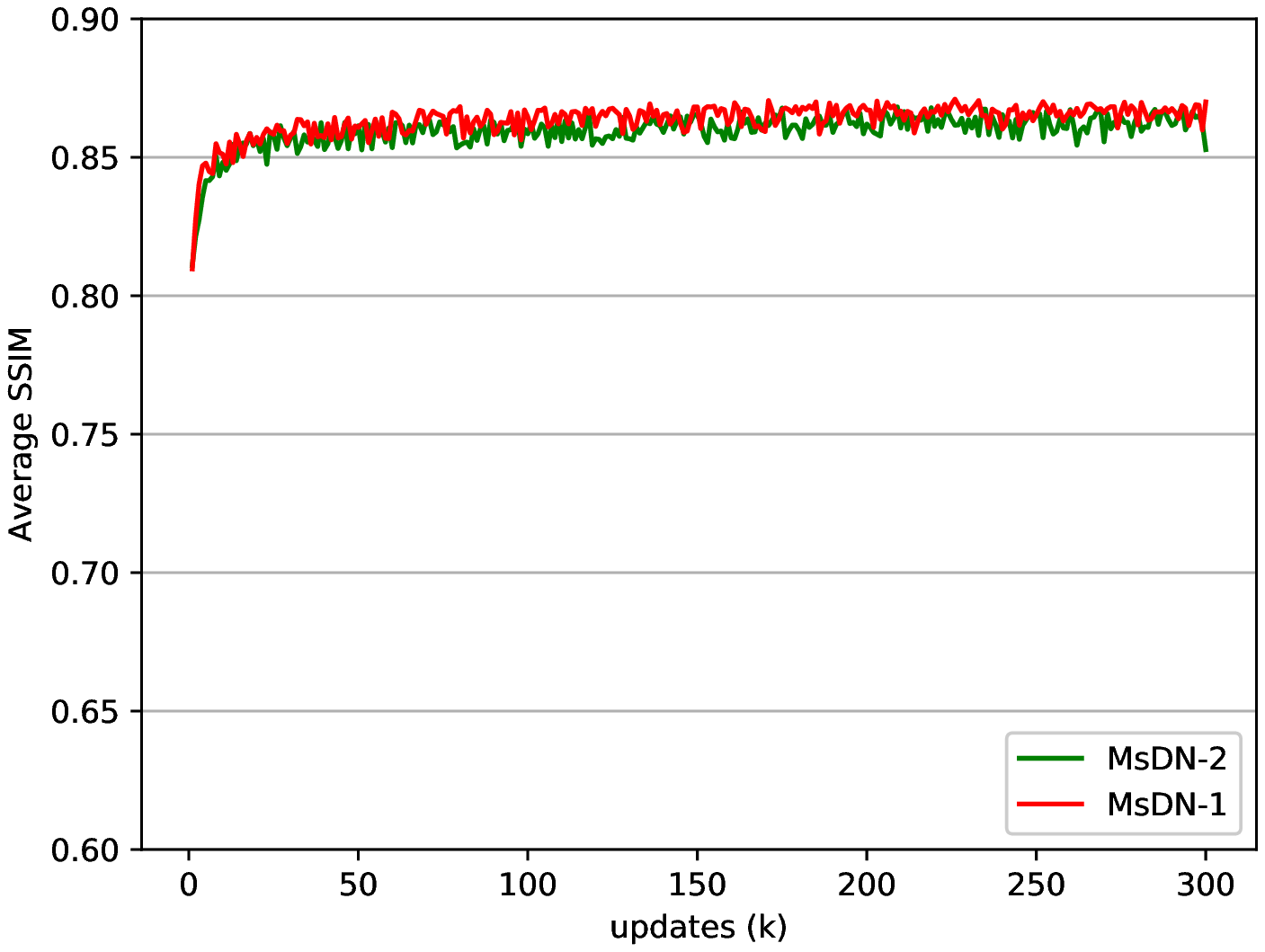}}
	\caption{The convergent curves of average PNSR and SSIM of MsDN on a cropped validation dataset in the training phase.}
	\label{fig:06}
\end{figure*}

Figure \ref{fig:05} shows the convergent curves of MsRN-1 and MsRN-2, from which we can see that the convergent speed of MsRN-1 is faster, and the average PSNR and SSIM is higher. Figure \ref{fig:06} shows the convergent curves of MsDN-1 and MsDN-2. The performance of MsDN-1 is more robust since its average PSNR and SSIM are higher. Therefore, according to the convergent curves of MsRN and MsDN, we conclude that $ L_1 $ norm minimization could be better in the training of MsDNN. 
\subsection{Comparisons of MsRN and MsDN}\label{sub:5.3}
To solve the image SR challenge from NTIRE 2019, we have developed two multi-scale networks, i.e., MsRN and MsDN. The architecture of MsRN is composed of the residual blocks while the development of MsDN is based on the dense blocks. To compare MsRN with MsDN quantitatively, we evaluate the performance of them on the validation dataset. As stated in Section \ref{sub:5.2}, we obtain four models by optimizing $ L_1 $ and $ L_2 $ losses, i.e., MsRN-2, MsDN-2, MsRN-1, and MsDN-1. Besides, the baselines of MsRN and MsDN are trained by $ L_1 $ norm to obtain two models, i.e., Baseline-R and Baseline-D. Finally, we compute the average PSNR and SSIM of each model on the validation dataset via the strategy in Section \ref{sec:4}. 
\begin{table*}[htp]
	\centering
	\caption{The average PSNR and SSIM of compared methods on the validation dataset. The blue value denotes the maximum among the values obtained by $ L_2 $ norm based models. The red value denotes the maximum among the values obtained by $ L_1 $ norm based models. }
	\label{tab:02}
	\setlength{\tabcolsep}{3.9mm}{
	\begin{tabular}{|c||c||c|c|c||c|c|c|}
		\hline 
		Method&  \multirow{2}{*}{Interpolation}&  \multirow{2}{*}{Baseline-R}& \multicolumn{2}{c||}{MsRN}&   \multirow{2}{*}{Baseline-D}& \multicolumn{2}{c|}{MsDN} \\ 
		\cline{1-1}\cline{4-5}\cline{7-8}
		Model&  &  &MsRN-2&  MsRN-1&  &MsDN-2&  MsDN-1\\ 
		\hline 
		PSNR&  27.78&  29.09&  29.42&  29.49&  29.02&  \color{blue}{29.47}&  \color{red}{29.51}\\ 
		\hline 
		SSIM&  0.8163&  0.8557&  0.8646&  0.8664&  0.8528&  \color{blue}{0.8651}&  \color{red}{0.8671}\\ 
		\hline 
	\end{tabular} }
\end{table*}

The quantitative evaluation of the models is showed in Table \ref{tab:02}, which shows that both PSNR and SSIM of MsDN are higher than them of MsRN. Figure \ref{fig:07} shows the visualization of SR images obtained by the models, which demonstrates that PSNR and SSIM of MsDN are comparable with them of MsDN. The average runtime of MsRN and MsDN on validation dataset are about 21 and 20 seconds, respectively. Therefore, we conclude that MsDNN is an effective method to solve the SR challenge from NTIRE 2019 according to our baselines, and the performance of MsRN and MsDN are comparable for single image SR.

\section{Discussion}\label{sec:6}
Utilizing the empirical experiments in Section \ref{sec:5}, we have showed the robustness of MsDNN for image SR with unknown upscaling factors. To test the performance of MsDNN for image SR with different downscaling operators, i.e., LR images are obtained by different cameras, we divide the validation dataset into two parts, i.e., \textit{Validation-1} and \textit{Validation-2}. the former is composed of the LR- and HR image pairs obtained by camera-1, the latter consists of the LR- and HR image pairs obtained by camera-2. Then we compute the average PSNR and SSIM of every model on \textit{Validation-1} and \textit{Validation-2}, as shown in Table \ref{tab:03}. The quantitative evaluation shows that the performance of MsDNN on \textit{Validation-2} is better than it on \textit{Validation-1}. Besides, we test the performance of models on Set5 \cite{set5}, Set14 \cite{set14} and DIV2K \cite{ntire01}. To be specific, each LR image is the biucbic interpolation ($ \times 3 $) of corresponding down-sampled HR image. Table \ref{tab:02a}  shows that the performance of MsDNN on the datasets is even worse than it of bicubic, of which the reason could be that there does not contain bicubic downscaling ($ \times 3 $) operator from HR to LR  image pairs in the training dataset of NTIRE 2019. Therefore, for the LR images obtained by different downscaling operators, the exploration of improving the generalization capability of DNN is still a difficult but interesting task.
\begin{table*}[htp]
	\centering
	\caption{The average PSNR and SSIM of compared methods on the datasets obtained by two different cameras. The blue value denotes the maximum among the scores of four models tested on Validation-1. The red value denotes the maximum among the scores of four models tested on Validation-2.}
	\label{tab:03}
	\setlength{\tabcolsep}{3.5mm}{
	\begin{tabular}{|c||c|c|c|c||c|c|c|c|}
		\hline 
		Dataset&  \multicolumn{4}{c||}{Validation-1}&  \multicolumn{4}{c|}{Validation-2}\\ 
		\hline 
		Model&  MsRN-2&  MsRN-1&  MsDN-2&  MsDN-1&  MsRN-2&  MsRN-1&  MsDN-2&  MsDN-1\\ 
		\hline 
		PSNR&  30.08&  30.16&  30.11&  \color{blue}{30.27}&  28.77&  28.82&  \color{red}{28.83}&  28.76\\ 
		\hline 
		SSIM&  0.8638&  0.8661&  0.8639&  \color{blue}{0.8677}&  0.8654&  \color{red}{0.8666}&  0.8662&  0.8665\\ 
		\hline 
	\end{tabular} }
\end{table*}
\begin{table*}[htp]
	\centering
	\caption{The average PSNR and SSIM of several models on Set5, Set14 and DIV2K under bicubic downscaling ($ \times 3 $) assumption. The red value indicates the best performance.}
	\label{tab:02a}
	\setlength{\tabcolsep}{3.26mm}{
		\begin{tabular}{|c||c||c||c|c|c||c|c|c|}
			\hline 
			\multirow{2}{*}{Dataset}& \multirow{2}{*}{Criteria}&  \multirow{2}{*}{Bicubic}&  \multirow{2}{*}{Baseline-R}& \multicolumn{2}{c||}{MsRN}&   \multirow{2}{*}{Baseline-D}& \multicolumn{2}{c|}{MsDN} \\ 
			\cline{5-6}\cline{8-9}
			&   &  &  &MsRN-2&  MsRN-1&  &MsDN-2&  MsDN-1\\ 
			\hline 
			\multirow{2}{*}{Set5}&PSNR&  \color{red}{28.63}&  27.38&  26.36&  26.56&  27.12&  26.83&  26.22\\ 
			&SSIM&  \color{red}{0.8378} &  0.8242&  0.8059&  0.8058&  0.8131&  0.8124&  0.8025\\ 
			\hline 
			\multirow{2}{*}{Set14}&PSNR&  \color{red}{25.89} &  25.20&  24.54&  24.44&  24.96&  24.60&  24.43\\ 
			&SSIM&  \color{red}{0.7401}&  0.7315&  0.7232&  0.7216&  0.7166&  0.7190&  0.7172\\ 
			\hline 
			\multirow{2}{*}{DIV2K}&PSNR&  \color{red}{29.19}&  28.12&  27.17&  27.05&  27.72&  27.33&  27.11\\ 
			&SSIM&  \color{red}{0.8359}&  0.8250&  0.8136&  0.8098&  0.8117&  0.8130&  0.8108\\ 
			\hline 
	\end{tabular} }
\end{table*}

\section{Conclusion}\label{sec:7}
The NTIRE 2019 image SR challenge is difficult since the upscaling factor of each LR image is unknown and the downscaling operators from HR to LR image pairs are different. Due to the fact that the sizes of an LR image and its ground truth are the same, processing the input images directly in LR space is extremely expensive. Besides, in the testing phase, the maximal number of filters of a convolutional layer is limited by the single GPU memory. For example, the number of filters of each convolutional layer must be less than 32 for a GPU with 11 GB memory in our experiments. To tackle these difficulties, we have proposed a multi-scale deep neural network to solve the SR challenge from NTIRE 2019. First, we have processed the input images mainly in the two downscaling spaces to simplify the computation complexity in the training and testing phases. Second, we have developed two deep neural networks in the downscaling spaces to super-resolve the LR images with unknown upscaling factors. Finally, we have tested the performance of MsDNN in Section \ref{sec:5} to show its robustness for image SR with unknown upscaling factors.

{\small
\bibliographystyle{ieee_fullname}
\bibliography{egbib}
}

\end{document}